%% file: main.tex
\documentclass[11pt,paper=a4]{scrartcl}
\usepackage[utf8]{inputenc}
\usepackage[T1]{fontenc}
\usepackage[margin=2.2cm]{geometry} 
\usepackage{amsmath,amssymb}
\usepackage{caption}
\usepackage{xcolor}
\definecolor{ScientificBlue}{RGB}{0, 0, 120}
\usepackage{hyperref}
\hypersetup{
    colorlinks=true,    
    urlcolor=ScientificBlue,      
    linkcolor=ScientificBlue,      
    citecolor=blue     
}
\usepackage{newtxtext} 
\usepackage{microtype} 
\usepackage{mdframed} 
\usepackage{graphicx}
\usepackage{caption}
\usepackage{enumitem}
\usepackage{orcidlink}
\usepackage{booktabs}
\usepackage{multirow}
\usepackage{siunitx}
\addtokomafont{title}{\color{ScientificBlue}}
\setkomafont{section}{\large\color{ScientificBlue}}
\RedeclareSectionCommand[
  afterskip=0.5em
]{section}
\setkomafont{subsection}{\color{ScientificBlue}} 
\RedeclareSectionCommand[
  afterskip=0.5em,
  beforeskip=-1.0em
]{subsection}
\setkomafont{subsubsection}{\small\color{ScientificBlue}} 
\RedeclareSectionCommand[
  afterskip=0.5em,
  beforeskip=-1.0em
]{subsubsection}
\addtokomafont{author}{\large}

\title{\LARGE A General Ambiguity Model for Binary Edge Images\\ with Edge Tracing and its Implementation}
\newcommand{\spacedcdot}{\hspace{0.3em}$\cdot$\hspace{0.3em}\ }
\author{Markus Hennig\,\orcidlink{0009-0000-1827-311X} \spacedcdot Marc Leineke \spacedcdot Bärbel Mertsching\,\orcidlink{0000-0003-0969-5243}}
\date{} 

\usepackage{algorithm} 
\usepackage[italicComments=false]{algpseudocodex} 
\usetikzlibrary{backgrounds}

\definecolor{color1}{rgb}{0.0, 0.45, 0.698} 
\definecolor{color2}{rgb}{0.0, 0.62, 0.45} 
\definecolor{color3}{rgb}{0.8, 0.47, 0.65} 
\definecolor{color4}{rgb}{0.84, 0.37, 0.0} 
\definecolor{color5}{rgb}{0.3, 0.3, 0.3} 
\newcommand{\colorA}[1]{{\color{color2}#1}}
\newcommand{\colorB}[1]{{\color{color2}#1}}
\newcommand{\colorC}[1]{{\color{color2}#1}}

\algrenewcommand\alglinenumber[1]{\footnotesize #1:}
\DeclareCaptionFont{sansserif}{\sffamily}
\DeclareCaptionFont{SB}{\color{ScientificBlue}} 
\newcommand{\subref}[1]{\textbf{\sffamily \textcolor{ScientificBlue}{#1}}} 
\newcommand{\sub}[1]{\textcolor{ScientificBlue}{#1}}
\usepackage[backend=biber, style=numeric, maxnames=3, minnames=3, giveninits=true]{biblatex}
\AtBeginBibliography{\small}

\DeclareNameAlias{sortname}{family-given}
\DeclareNameAlias{default}{family-given}

\DeclareFieldFormat{doi}{%
    DOI\addcolon\space
    \ifhyperref
      {\href{https://doi.org/#1}{\nolinkurl{#1}}}
      {\nolinkurl{#1}}
}

\renewbibmacro{in:}{}

\DeclareFieldFormat{year}{\mkbibparens{#1}}

\AtEveryBibitem{
  \clearfield{month}%
  \clearfield{day}%
  \clearfield{issn}%
}

\usepackage[capitalize]{cleveref}

\Crefname{figure}{Figure}{Figures}
\usepackage{afterpage}

\begin{document}

\maketitle

\vspace{-1.7cm}
\begin{center}
\small{GET Lab, Department of Electrical Engineering and Information Technology, Paderborn University, Germany}

Corresponding author: \small{\texttt{%
\href{mailto:hennig@get.upb.de}{hennig@get.upb.de}}}%
\end{center}

\input{abstract}

\input{1-introduction}
\input{2-related-work}
\input{3-tracing}
\input{4-evaluation}
\input{5-conclusion}


\small
\printbibliography
\normalsize

\end{document}

%% file: abstract.tex
\definecolor{vlgray}{gray}{0.95}
\begin{mdframed}[backgroundcolor=vlgray, linecolor=white, innertopmargin=10pt, innerbottommargin=10pt]
\textsf{\textcolor{ScientificBlue}{\textbf{Abstract.}}}
We present a general and intuitive ambiguity model for intersections, junctions and other structures in binary edge images. The model is combined with edge tracing, where edges are \textit{ordered} sequences of connected pixels. The objective is to provide a versatile preprocessing method for tasks such as figure-ground segmentation, object recognition, topological analysis, etc. By using only a small set of straightforward principles, the results are intuitive to describe. This helps to implement subsequent processing steps, such as resolving ambiguous edge connections at junctions. By using an augmented edge map, neighboring edges can be directly accessed using quick local search operations. The edge tracing uses recursion, which leads to compact programming code. We explain our algorithm using pseudocode, compare it with related methods, and show how simple modular postprocessing steps can be used to optimize the results. The complete algorithm, including all data structures, requires less than 50 lines of pseudocode. We also provide a C$++$ implementation of our method.
\end{mdframed}


\textsf{\small \textbf{Implementation:}} {\small \url{https://github.com/mh-upb/edge-tracing}}

%% file: 1-introduction.tex
\section{Introduction}
\label{sec:introduction}
In this work, we present a general model for ambiguities in binary edge images $f: \mathbb{Z} \times \mathbb{Z} \rightarrow \{0, 1\}$. We define ambiguities as pixels where the paths of edges are not clearly defined, such as at intersections or junctions (a formal definition is given in \cref{sec:proof}). Our method is combined with edge tracing, where the objective is to read in edge pixels in an ordered sequence. An edge is defined as a chain of connected pixels labeled 1, and the order is determined by following along the edge. The idea is to provide an intuitive and flexible ambiguity model for subsequent processing steps, such as connecting edge segments, extracting coherent object contours, or bridging gaps. We focus on binary edge images with mostly one-pixel wide edges, as computed by many edge detection algorithms (where binarization is typically a post-processing step) \cite{jing2022,yang2022}. Note that the term edge tracing or tracking is also used in the context of \textit{edge detection} in real images, when gradients and other image structures are traced to determine coherent edge curves, such as in \cite{burke2022}. 

Edge images are interesting to analyze because they provide compact yet descriptive scene representations. They can be used to extract object contours and other salient structures, which are important features for many higher-level vision tasks \cite{li2015,yang2017}. Edges and contours are also important features in human vision \cite{walther2011,singh2015}. In digital images, edges can be detected even in case of challenging illumination conditions and complex textures, especially when using learning instead of traditional gradient-based methods \cite{jing2022,yang2022}. As these methods are continuously improved, it is also interesting to develop new methods to process the resulting edge images.

Coherent object contours in the form of traced edges (i.e., ordered sequences of connected edge pixels) are required for many standard shape description methods such as Fourier descriptors, chain codes, signature functions, and Curvature Scale Space analysis \cite{mokhtarian2003,wang2018}. Another potential application is the generation of object bounding box proposals, such as in \cite{zitnick2014}. Additionally, our model could be used in applications where explicit edge connection information is of interest, such as determining cell–cell contact phenotypes, as in \cite{brezovjakova2019}, or describing vessel crossings, as in \cite{wang2023}.

\begin{figure}[t]
\centering %
\vspace{0.5cm}
\includegraphics{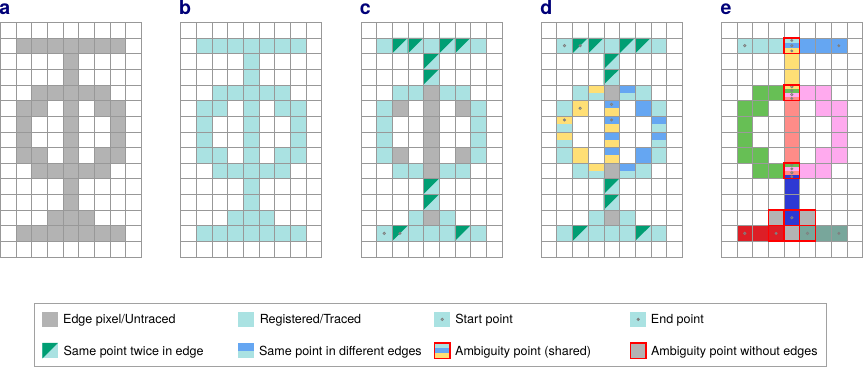}
\parbox{\textwidth}{\raggedright
{\scriptsize
\vspace{-3.85cm}
\hspace{1.45cm} Input image
\hspace{1.0cm} CCL \cite{he2017} (unordered)
\hspace{0.6cm} MNT \cite{ghuneim2000} (ordered)
\hspace{0.75cm} FCM \cite{suzuki1985} (ordered)
\hspace{0.95cm} Ours (ordered)}}
\vspace{-0.75cm}
\caption{Comparison of different standard methods for processing binary edge images with ours. \subref{a} Input image. \subref{b}--\subref{e} Results of the different methods. Different colors represent different edges. Our model provides direct access to ambiguities, registers all pixels, and avoids redundancies.}
\label{fig:comparison}
\end{figure}

\cref{fig:comparison} shows the results of common methods for processing binary edge images compared to ours. The methods are described in \cref{sec:related-work}. In summary, our method registers \textit{all} pixels without redundancy (no double reading per edge), provides a structured decomposition into meaningful edges, and direct access to ambiguities. In comparison, Connected Component Labeling (CCL) does not provide any order, while Moore-Neighbor Tracing (MNT) and the Find Contours Method (as named in OpenCV, FCM) miss some inner pixels, including those in the vertical line running through the circle. Furthermore, these methods do not directly model ambiguities of edge paths.

In summary, we present an ambiguity model that is intuitive, flexible, and more general than existing models. We provide a compact pseudocode version of our algorithm to describe its workflow and facilitate its reimplementation in different programming languages. Additionally, we provide a C++ implementation of our algorithm as open-source software. Furthermore, we present selected application examples to demonstrate the versatility of our method. Finally, we compare our method with other standard methods for processing binary edge images using different edge detectors on the BSDS500 dataset \cite{arbelaez2011}, providing insights on component decomposition, redundancy, and missing pixels.

%% file: 2-related-work.tex
\section{Related Work}
\label{sec:related-work}
One of the most common methods for processing binary edge images is Connected Component Labeling (CCL) \cite{he2017}, which simply labels connected pixels. MNT \cite{ghuneim2000} is a standard contour tracing method and, in this work, serves as a representative for many similar methods with slight modifications. A comprehensive comparison can be found in \cite{seo2016}. Besides MNT, the authors review existing contour tracing methods, such as the Square Tracing Algorithm, Radial Sweep, and Theo Pavlidis' algorithm. FCM \cite{suzuki1985} is another popular tracing method that also detects hierarchies between edges (parents and childs). However, none of the methods provide an ambiguity model. Since MNT and FCM are region contour tracers, they often trace edge pixels more than once per edge, leading to redundancies. In contrast, our method traces each edge pixel only once per edge. Some inner pixels are not traced at all in MNT and FCM.

Compared to existing works, our method provides direct access to ambiguities and connected traced edges in the image plane, relating them without complex multi-stage processes or graph representations. Our method \textit{describes} binary edge images and provides an intermediate representation independent of any specific edge detection method or application-specific objectives and metrics. The work in \cite{heng2001} focuses on identifying specific simple edge configurations for splitting edges, while \cite{law1996} explores different types of joins, including scenarios with three edges at a triple point and two edges at a junction. The method in \cite{guo2014} consists of six stages and groups edge segments based on cues from real images. Different from our model, ambiguities are considered as options for connecting mostly free-standing edge segments, are modeled using graphs, and the overall results are evaluated in terms of edge \textit{detection} results. The method in \cite{pham2014} detects junctions in binary line drawings by searching for optimal meeting points and classifies them into different types. Unlike our method, it does not model multi-pixel ambiguities, a larger number of edges connected to single ambiguities, or provide direct access to each segment.
The method in \cite{casadei1999} is a multi-stage algorithm using a contour graph to group edge segments, where the concept of ambiguities is not as straightforward as ours. The method in \cite{kimia2019} models ambiguities as possible edge connections based on propagated curve bundles, which is also more complex. The method in \cite{buades2018} detects line segments and other edge image elements, and the method in \cite{huang2018} detects junctions in real images to extract wireframe representations of man-made environments, but both methods do not provide an ambiguity model. The method in \cite{tamrakar2007}, a preliminary work to \cite{kimia2019}, represents edge segment combinations using curve bundles, and the ambiguity model is more complex and has different objectives than ours. The method in \cite{zhu2007} groups edge segments to search for cyclic structures using a directed graph, and the method in \cite{maire2008} combines edge detection with junction detection in real images, but both methods do not provide an ambiguity model. Other methods aim to accelerate the contour tracing, as explored in \cite{gupta2022}. In summary, most existing works focus on grouping edge segments to refine edge detection results, rather than directly describing these results in a flexible and intuitive manner. Our method could likely be integrated as an intermediate step in such works.

%% file: 3-tracing.tex
\section{Ambiguity Model and Tracing}
\label{sec:main-model}
Ambiguities in binary edge images occur when edges are more than one pixel wide, when edges cross or meet at intersections or junctions, or when the edge detection process introduces artifacts. As a result, edges can share one or more pixels and they can be adjacent to or running through pixel clusters. When edges meet at a single pixel, such as at T-, Y-, and X-junctions, as shown in Figure~\ref{fig:test-figures}\sub{b}, they only share that pixel. However, more complex junctions can extend over several pixels (pixel clusters) with variable shapes and sizes, as shown in \cref{fig:test-figures,fig:examples}, which cannot be effectively processed by implementing a separate case for each variation.

For some conceptual descriptions, we distinguish between the two types as single-pixel ambiguities (SPAs) and multi-pixel ambiguities (MPAs). For our model and implementation, this explicit distinction is not required, and one can simply check the size (number of corresponding pixels) of an ambiguity if necessary. Depending on the context, we refer to pixels as points, especially when explicitly considering their spatial coordinates $(x,y)$.



\subsection{Modeling Principles}
\label{sec:principles}
As already outlined, edges can share single pixels (SPAs) and they can be connected to pixel clusters (MPAs). Our model integrates these cases in a general and intuitive way, without the need to distinguish the two types or various specific pixel configurations. It uses the following straightforward principles, which work in conjunction with each other:
\begin{enumerate}[itemsep=0pt, parsep=0pt]
    \item Edges adjacent to ambiguities are connected to them via the shortest path.\label{pr1}
    \item The connection pixel is the start or end point of the respective edge.\label{pr2}
    \item Each pixel cluster forms a single coherent ambiguity.\label{pr3}
    \item Cluster pixels without edge connections are preserved but not traced.\label{pr4}
\end{enumerate}
%
%
We have derived these principles from extensive tests with real and artificial binary edge images to develop an effective model. Principle \ref{pr1} is inherently satisfied for SPAs, as there is only one direct connection option. For MPAs, it is consistent with edge tracing, where adjacent edge pixels are also connected via the shortest path. Principle \ref{pr2} is inherently satisfied for SPAs, since all edges share the respective pixel, as shown in Figure~\ref{fig:test-figures}\sub{b}. It naturally extends to MPAs by integrating the connection pixel into each edge as well. Having only a single connection pixel is meaningful because edge paths within ambiguities are not clearly defined. From an implementation perspective, the principle provides a clear link between edges and ambiguities, which is helpful to determine potential edge connections. Principle \ref{pr3} maintains clarity despite variable shapes and sizes of clusters. By treating each cluster as a single coherent ambiguity, all connected edges are linked by a single entity, just as with SPAs. Principle \ref{pr4} can only take effect for MPAs (clusters), as SPAs always have edge connections by definition. It ensures that all pixels leading to an ambiguity are preserved and that the untraced pixels within an ambiguity are directly accessible. This is interesting to determine potential edge connections that may exclusively use detected edge pixels. In summary, Principles \ref{pr1}--\ref{pr3} can be seen as natural generalizations from SPAs to MPAs, and Principle \ref{pr4} as a logical consequence of unclear edge paths.

\begin{figure}[t]
\centering %
\includegraphics{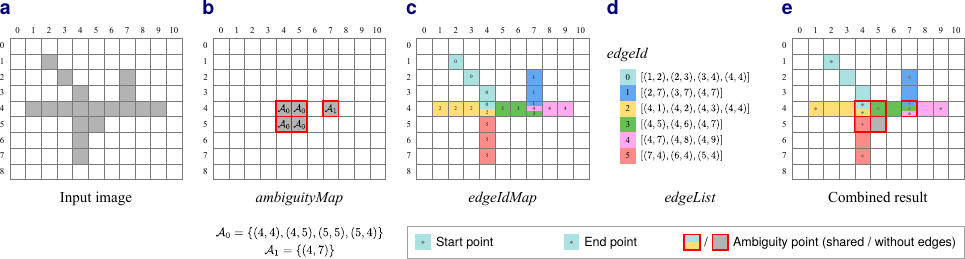}
\caption{Modeling idea and central data structures of our algorithm. \subref{a} Binary edge image with two ambiguities. \subref{b},\ \subref{c} Together, the \textit{ambiguityMap} and \textit{edgeIdMap} form an augmented edge map. \subref{d} The traced edges are stored in the \textit{edgeList}. \subref{e} Combined result with shared pixels and connections to ambiguities.}
\label{fig:maps}
\end{figure}

\subsection{Procedure and Pseudocode Implementation}
\label{sec:procedure-and-pseudocode}
Our algorithm operates in two main passes: First, all ambiguities are identified so that the remaining pixels after this step are exclusively edge pixels to be traced. Second, these edge pixels are traced in sequential order, during which the resulting edges are connected to adjacent ambiguities, if present. This two-pass structure provides clear operational control and helps to verify that our algorithm works as intended.

In the following pseudocode implementation, \cref{alg1} controls the overall operation, which is divided into the two main passes mentioned (preprocessAmbiguities in \cref{alg1:callPre} and the for-loop beginning at \cref{alg1:forLoop}). The pseudocode conceptually aligns with our C++ implementation.

\subsubsection{Data Structures}
\label{sec:data-structures}
As shown in \cref{fig:maps}, our algorithm utilizes three central data structures: \textit{ambiguityMap}, \textit{edgeIdMap}, and \textit{edgeList}. Both the \textit{ambiguityMap} and \textit{edgeIdMap} have the size of the original input image and jointly form an augmented edge map.

The \textit{ambiguityMap} stores all ambiguity points, where each point stores the coordinates of \textit{all} pixels belonging to the respective ambiguity $\mathcal{A}_i$. Thus, every point belonging to the respective ambiguity can be directly retrieved from every point in that ambiguity. 

The \textit{edgeList} holds all traced edges, where an \textit{edge} is an ordered sequence of connected edge pixels. The position of each \textit{edge} in this list is also its identifier, a simple integer referred to as \textit{edgeId}.

The \textit{edgeIdMap} stores, at each point, a list containing the \textit{edgeId}s of every \textit{edge} running through that point (the list is empty if no edges are running through). Thus, neighboring edges (and their data) can be directly accessed using quick local search operations in the \textit{edgeIdMap}.

Together, the \textit{ambiguityMap} and \textit{edgeIdMap} provide direct access to every edge connected to an ambiguity. Without the \textit{ambiguityMap}, it is not directly clear which edges are connected to specific ambiguities.
Other edge-processing tasks, such as bridging gaps between nearby edges, can also be addressed in an efficient manner.

\subsubsection{Core Subfunctions}
\label{sec:core-subfunctions}
\begin{figure}[t]
\centering %
\includegraphics{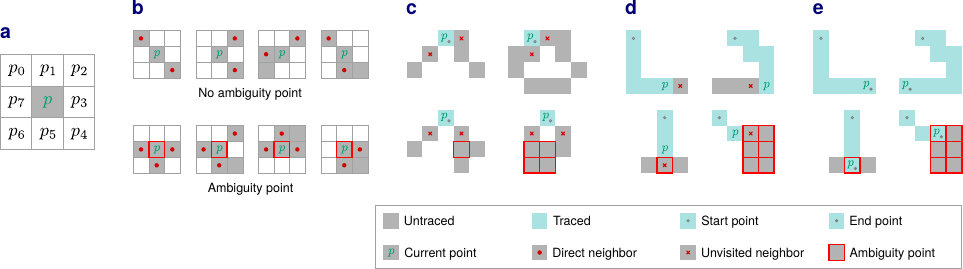}
\caption{Building blocks of our algorithm. \subref{a} Naming of the 8-neighbors of the current point. \subref{b} Mechanism for identifying ambiguity points. \subref{c}--\subref{e} Tracing with two, one, and no unvisited neighbors, respectively.}
\label{fig:cases}
\end{figure}

Our algorithm utilizes three straightforward core subfunctions: getDirectNeighbors, containsFourCluster, and mergeEdges. The first two functions are used in conjunction to check if the current point is part of an ambiguity, as in \cref{alg2:isCluster1,alg2:isCluster2} of \cref{alg2}. Additionally, the function getDirectNeighbors is used in \cref{alg3} in \cref{alg3:getDirectNeighbors} to trace (connect) adjacent edge pixels via the shortest path. The function mergeEdges is used in \cref{alg3} to combine two edges into a single edge during the tracing. The naming of the 8-neighbors of the current point $p$ for the following descriptions is shown in \cref{fig:cases}\sub{a}.

The function getDirectNeighbors analyzes the 8-neighbors of the current point $p$. The function operates in a specific way and does not simply return \textit{all} neighbors that are set, as shown with examples in \cref{fig:cases}\sub{b}: It returns all orthogonal neighbors ($p_1, p_3, p_5, p_7$) that are set, but from the diagonal neighbors ($p_0, p_2, p_4, p_6$) \textit{only those} which do not have any set orthogonal neighbor in ($p_1, p_3, p_5, p_7$). We denote the returned neighbors as \textit{direct neighbors}. If the function returns more than two direct neighbors, the current point is definitely part of an ambiguity. Note that this check is not sufficient to identify all ambiguities; the result of the function containsFourCluster must also be considered (cf.\ examples in \cref{fig:cases}\sub{b}).

The function containsFourCluster checks if the current point $p$ is located within a four-cluster (a $\text{2}\times\text{2}$ block of set pixels). In this case, every point in at least one of the groups ($p_7, p_0$, $p_1$), ($p_1, p_2$, $p_3$), ($p_3, p_4$, $p_5$), or ($p_5, p_6$, $p_7$) is set. This check can be implemented in a straightforward manner by encoding the occupancy of $p_0$--$p_7$ in a binary number and checking if it contains the respective cases. If the current point is located within a four-cluster, it is definitely part of an ambiguity.

The function mergeEdges combines two edges into one and updates the \textit{edgeIdMap} accordingly. This is required when the tracing starts at a point within an edge, where it initially runs in two directions and the resulting edges are finally merged. It can also be used during postprocessing, such as when connecting edges in ambiguities (see \cref{sec:application-examples}). The function operates on the \textit{edgeList} based on two passed \textit{edgeIds}. The merged edge retains the smaller of the two \textit{edgeIds} (which is technically not necessary, but provides a clear system). The two edges must share an overlapping point at one of their sides, serving as the connection point. The function ensures the correct order of the traced points by considering the position of the overlapping point. There are four cases to consider: both edges start at the same point, both edges end at the same point, the first edge starts where the second ends, or the first edge ends where the second starts.

\subsubsection{Main Function (Algorithm \ref{alg1})}
\label{sec:main-function}
The main function in \cref{alg1} controls the overall operation. In \cref{alg1:callInit}, the \textit{ambiguityMap} and \textit{edgeIdMap} are initialized as empty data structures with the size of the input image, and the \textit{edgeList} as an empty list. These data structures are modified by the functions preprocessAmbiguities and traceEdge. In \cref{alg1:callPre}, the function preprocessAmbiguities identifies all ambiguities and stores them in the \textit{ambiguityMap} (see \cref{sec:preprocessing} for details). After that, the construction of the \textit{ambiguityMap} is finished (see \cref{fig:maps}\sub{b} for an example), but no edges have been traced yet. This is done in the subsequent for-loop beginning at \cref{alg1:forLoop}. The if-statement in \cref{alg1:ifCheck} checks if the current point is an edge pixel, is not part of an ambiguity, and has not been traced yet. If these conditions are met, a new \textit{edge} is created as an empty PointList. This list is then passed to the function traceEdge, along with the input image and the current point, initiating the tracing of the edge to which that point belongs (see \cref{sec:edge-tracing} for details).

\afterpage{%
\clearpage 
\vspace*{0.2cm}
\input{3a-pseudocode}
\input{3b-pseudocode}
\input{3c-pseudocode}
\clearpage 
}%

\subsubsection{Preprocessing (Algorithm \ref{alg2})}
\label{sec:preprocessing}
The function preprocessAmbiguities in \cref{alg2} implements the first pass of our algorithm. In summary, once a point is identified as part of an ambiguity, all direct neighbors with the same property are iteratively registered until all points belonging to the current ambiguity are included (corresponding to modeling Principle~\ref{pr4}).
The while-loop beginning at \cref{alg2:while} checks for each direct neighbor $p_n$ if it is also part of the current ambiguity. Such points are iteratively appended to the list. If the passed point $p$ is just a SPA, no further points are appended. Comparing the counter variable $c$ with the size of the list in \cref{alg2:while} ensures that the direct neighbors of each appended point are also checked, similar to a region-growing process.

\subsubsection{Edge Tracing (Algorithm \ref{alg3})}
\label{sec:edge-tracing}
This function traces the edge to which the passed point $p$ belongs and connects the edge to any ambiguities on its sides, if present.
The check in \cref{alg3:part1Start} ensures that the tracing does not continue within an ambiguity. The second check in line \ref{alg3:part1Ambiguity} ensures that a connection pixel from an adjacent ambiguity becomes part of the edge (corresponding to modeling Principle~\ref{pr2}; cf.\ \cref{fig:cases}\sub{d},\,\sub{e}, bottom row). After these steps, the \textit{unvisitedNeighbors} list contains either two, one, or no points.

Consider \cref{fig:cases}\sub{c}: In case of two points in the \textit{unvisitedNeighbors} list, which is covered in Lines \ref{alg3:start2Neighbors}--\ref{alg3:end2Neighbors}, the tracing of a new \textit{edge} has just started and the passed point $p$ is located somewhere within the \textit{edge}. In this case, two new edges, running in the directions defined by the two points in the list, are created and traced. The passed point $p$ serves as the starting point of both edges (as \textit{mergeEdges} requires an overlapping point). After the tracing of both edges is finished, the edges are merged in \cref{alg3:end2Neighbors}. The corresponding \textit{edgeIds} result from incrementing the \textit{edgeId} after the two edges have been completely traced.

Consider \cref{fig:cases}\sub{d}: In case of one point in the \textit{unvisitedNeighbors} list, which is covered in Lines \ref{alg3:start1Neighbor}--\ref{alg3:recur3}, the tracing continues in the direction of this point.

Consider \cref{fig:cases}\sub{e}: In case the \textit{unvisitedNeighbors} list is empty, which is covered in Lines \ref{alg3:start0Neighbors}--\ref{alg3:end0Neighbors}, the tracing of the current \textit{edge} is finished. In this case, the \textit{edge} is appended to the \textit{edgeList}, and the \textit{edgeId} is incremented.

\subsection{Proof of Correctness}
\label{sec:proof}
We consider our algorithm to be correct when it aligns with the modeling principles introduced in \cref{sec:principles}. The proof can be broken down into verifying its three core steps: that all ambiguities are correctly identified during the preprocessing, that the edge pixels remaining after the preprocessing are traced in the correct order, and that edges are correctly connected to adjacent ambiguities during the tracing.

As reflected in \cref{alg2:isCluster1,alg2:isCluster2} of \cref{alg2}, an ambiguity point is defined as the center of a $\text{3}\times\text{3}$ neighborhood that has more than two direct neighbors or is located within a four-cluster. The examples shown in \cref{fig:cases}\sub{b} can be easily extended and clearly demonstrate that this criterion is correct. For a systematic proof, one can verify all $2^8 = 256$ possible configurations of the 8-neighbors. As the same criterion is applied to all direct neighbors of an ambiguity point, all points belonging to a specific ambiguity are also correctly identified.

The remaining edge pixels are traced in the correct order because the tracing direction is determined by the points returned by the function getDirectNeighbors. This function does not return diagonal neighbors that have an orthogonal neighbor, thus giving priority to orthogonal connections. Since an orthogonal neighbor has a distance of 1 from the center, and a diagonal neighbor a distance of $\sqrt{2}$, the tracing follows the shortest and therefore correct path along the edge, as shown in \cref{fig:cases}\sub{b} (top row).

Since edges adjacent to ambiguities are connected to them using the principle from the edge tracing (following the shortest path), the connection step is also correct.

\subsection{Simplified Implementation}
\label{sec:simplified-impl}
Depending on the task, our implementation can also be simplified, particularly in terms of the data stored in the \textit{ambiguityMap} and \textit{edgeIdMap}. In its current form, the \textit{ambiguityMap} stores a list of points at each ambiguity point, providing direct access to every connected edge via the \textit{edgeIdMap}. If this access and checking the size of ambiguities are not required, the \textit{ambiguityMap} can be replaced with a simple binary map that only encodes the presence of ambiguity points. In this case, \cref{alg2:writeAmbiguityMapPoints} in \cref{alg2} can directly write binary values.

In a similar manner, the \textit{edgeIdMap} can be replaced with a binary map, if direct access to the edges passing a specific point and local searches for neighboring edges are not required. In this case, \cref{alg3:edgeIdMapAppend} in \cref{alg3} can also directly write binary values.

If the objective is to exclusively trace edges that start or end at ambiguities according to the principles, the \textit{ambiguityMap} and \textit{edgeIdMap} can be merged into a single binary map that encodes whether a point has already been processed. However, this requires some deeper modifications in the code.

\begin{figure}[t]
\centering %
\includegraphics{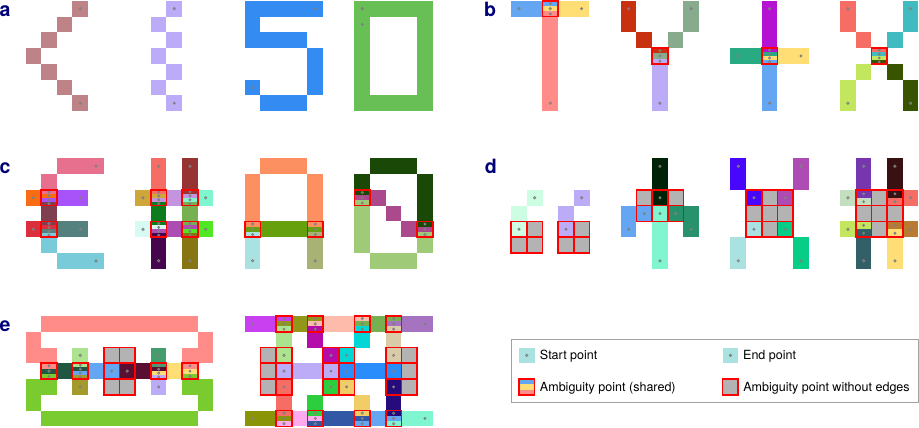}
\vspace{0.3cm}
\caption{Example results for some fundamental test cases, where different colors represent different edges. \subref{a} Tracing without ambiguities. \subref{b},\,\subref{c} Single-pixel ambiguities. \subref{d} Multi-pixel ambiguities. \subref{e} Complex nested cases.}
\label{fig:test-figures}
\end{figure}

\subsection{Fundamental Test Cases}
\label{sec:fundamental-test-cases}
\cref{fig:test-figures} shows the results of our algorithm for some fundamental test cases. In all cases, all pixels from the input image have been registered and are displayed, either as traced edge pixels or as ambiguity points. \cref{fig:test-figures}\sub{a} shows three open edges and one closed edge without any ambiguities, so the edges are simply traced in sequential order. \namecrefs{fig:test-figures} \ref{fig:test-figures}\sub{b} and \ref{fig:test-figures}\sub{c} show examples for SPAs, including T-, Y-, and X-junctions. \cref{fig:test-figures}\sub{d} shows examples for MPAs, and \cref{fig:test-figures}\sub{e} complex nested cases. In summary, our method works as intended and the corresponding principles provide a clear, intuitive and natural description of all cases despite their sometimes significant complexity.

\subsection{Application Examples}
\label{sec:application-examples}

\begin{figure}[t]
\centering %
\includegraphics{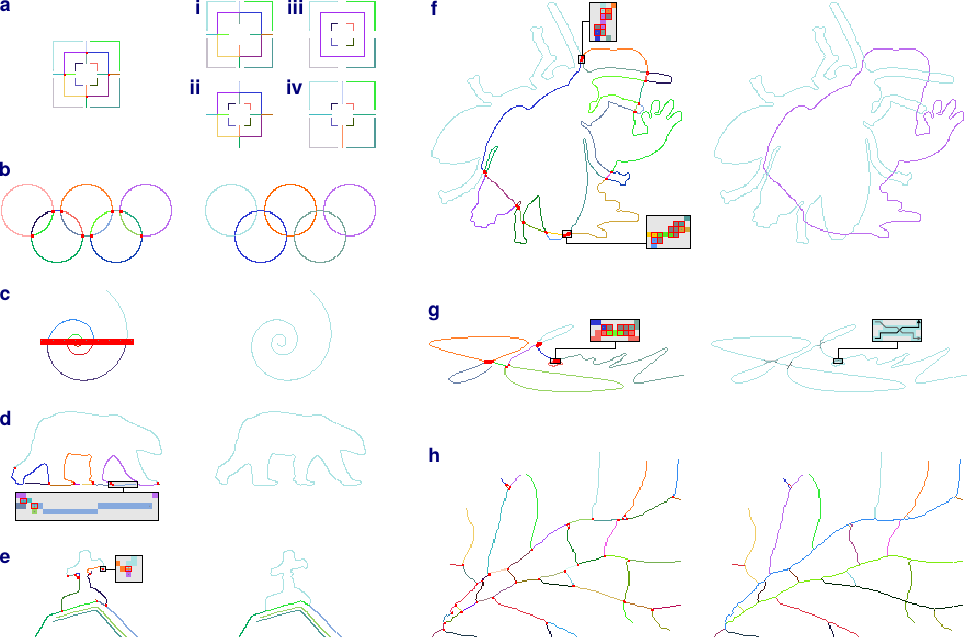}
\vspace{0.3cm}
\caption{Results for different examples and postprocessing steps (see text and zoom in for details). In each subfigure: Left: Initial model output. Right: Resulting edges. \subref{a} Four different postprocessing strategies based on the edge length and connection to ambiguities (free, dangling, bridged). \subref{b} The individual rings are separated. \subref{c} The spiral segments are connected. \subref{d} The bear is represented as a single, closed contour. \subref{e} The cross is extracted as a single contour. \subref{f} The two figures are separated despite complex ambiguities. \subref{g} The path of the pen stroke followed during the signature is traced (resulting path directions indicated by arrows). \subref{h} The main branches of the vessels are identified.}
\label{fig:examples}
\end{figure}

\cref{fig:examples} shows the results of our algorithm for selected application examples and postprocessing steps. Such steps can be combined and applied in different orders in a modular fashion, depending on the specific input and objectives. \namecrefs{fig:examples}~\ref{fig:examples}\sub{a}--\sub{c} and \ref{fig:examples}\sub{f} are artificial examples.

In \cref{fig:examples}\sub{a}, specific edges have been removed depending on their connection to ambiguities and length, with this information being directly accessible through our model: In Example \sub{i}, all free-standing edges without connections to ambiguities and shorter than 20 pixels have been removed. In Example \sub{ii}, all free-standing edges longer than 10 pixels have been removed. In Example \sub{iii}, all edges that have a connection to only one ambiguity (either at the start or end, dangling) have been removed. Note that this resolves all ambiguities in the middle edge, resulting in a single, closed edge. In Example \sub{iv}, all edges that have a connection to an ambiguity on both sides have been removed. Such postprocessing options could be helpful when short or long edges need to be disconnected from certain other edges, or to remove clutter from edge images.

In \cref{fig:examples}\sub{b}, the ambiguities have been resolved by connecting the respective edges using a simple cost function approach. Recall that each pixel cluster forms a single coherent ambiguity (Principle~\ref{pr3}), so each ambiguity has been processed separately. The cost function takes into account the angle difference at which two edges approach an ambiguity and the Euclidean distance between the connection points, with the objective of connecting edges based on good continuity. The angle relative to an ambiguity is computed using a simple least squares line fit based on the last $N$ pixels from the connection point (in this example, we have used 5 pixels, but the exact number is not critical). An edge is connected to another edge or itself if the cost for the candidates is the lowest, provided that the cost is below a certain threshold. This process continues until all edges connected an ambiguity have been checked. Note that this approach is simple and effective, but not necessarily optimal in terms of all possible edge connections, as we are directly taking the next best match. Edges are connected by straight lines between the connection points (which are start or end points), generated using the Bresenham line algorithm \cite{bresenham1965}. To improve the fit, a method that takes into account the edge paths could be employed, such as interpolation using Euler spirals \cite{connor2014}.

\cref{fig:examples}\sub{c} has also been processed using the cost function approach. However, in this example, all edges are connected to the same large ambiguity. In the cost function, the Euclidean distance between the connection points has been weighted higher than the angle difference (so that larger distances lead to higher costs) to prioritize the connection of opposite edges.

\cref{fig:examples}\sub{d} shows a region cropped from image 100007 in the BSDS dataset after applying the Canny edge detector. In this example, the bear cannot be directly analyzed as one coherent contour, as it is connected to some short edges caused by detection artifacts and some additional edges caused by shadows in the lower area. From the model output, all dangling edges shorter than 30 pixels (the exact number is not critical) have been removed twice in succession. Consider the magnified region in \cref{fig:examples}\sub{d}: Repeating this process twice is required because, after the first run, only the two dangling edges are removed, leading to one remaining dangling edge. In principle, such a process could be repeated until no more changes occur. Another strategy here and in general could be to analyze connections across multiple ambiguities to see if this results in closed contours, because it cannot be guaranteed that additional edges are always just dangling.

\cref{fig:examples}\sub{e} shows a region cropped from image 118035 in the BSDS dataset after applying the Canny edge detector. In this example, all edges shorter than 6 pixels have been removed, independent of their connections to ambiguities. The contour of the cross was initially connected to some short edges, which have been removed, as shown in the magnified region, but also to an edge with connections to ambiguities on both sides.

\cref{fig:examples}\sub{f} shows an artificial example with many complex ambiguities. As shown in the magnified regions, ambiguities can be easily located close to each other, and due to the several edges converging at such points, such ambiguities should be considered together. Therefore, our first postprocessing step has been to merge ambiguities that are connected by edges with a length of 3 pixels into one combined ambiguity, so that all connected edges can be accessed and connected (the corresponding function is provided with our implementation). Next, we have connected edges at ambiguities using the previous cost function approach. In summary, only two postprocessing steps have been required to separate the two figures. As a result, our model could be helpful for figure-ground segmentation and similar tasks.

\cref{fig:examples}\sub{f} shows an example obtained by half-automated binarization of image 39\_24 from the CEDAR signature dataset \cite{cedar2007}. The example has also been postprocessed using the cost function approach. Additionally, the order of the resulting edge has been reversed so that the start point corresponds to the start of the pen stroke. A characteristic difference from \cref{fig:examples}\sub{f} is that it is a single self-intersecting contour rather than two mutually intersecting contours.

\cref{fig:examples}\sub{h} shows a region cropped from image 21 in the DRIVE retina image dataset \cite{staal2004} after applying binarization and morphological skeletonization. In the first step, all edges shorter than 3 pixels have been removed. Next, the cost function approach from the previous examples has been used. The next step could be to convert the connections of main and sub-vessels into a graph representation to obtain a model with data for medical purposes.

%% file: 3a-pseudocode.tex
\begin{algorithm}[ht]
\caption{Main}
\begin{small}
\label{alg1}
\begin{algorithmic}[1] 
\State Initialize: \textit{ambiguityMap}, \textit{edgeIdMap}, \textit{edgeList}\label{alg1:callInit} \Comment{Global data structures modified by \cref{alg2,alg3}}
\Function{\textnormal{main}}{Image $I_\text{in}$}\label{alg1:callMain}
    \State preprocessAmbiguities($I_\text{in}$) \Comment{\cref{alg2}}\label{alg1:callPre}
    \For{\textbf{each} Point $\colorB{p}=(x,y)$ \textbf{in} $I_\text{in}$}\label{alg1:forLoop}
        \If{$I_\text{in}$[$\colorB{p}$] $>$ 0 \textbf{and} \textit{ambiguityMap}[$\colorB{p}$].size $==$ 0 \textbf{and} \textit{edgeIdMap}[$\colorB{p}$].size $==$ 0} \label{alg1:ifCheck}
            \State \textbf{new} PointList \textit{\colorC{edge}} \Comment{New empty \textit{edge}}
            \State traceEdge($I_\text{in}$, $\colorB{p}$, \textit{\colorC{edge}}) \Comment{\cref{alg3}}
        \EndIf
    \EndFor
\EndFunction
\end{algorithmic}
\end{small}
\end{algorithm}

%% file: 3b-pseudocode.tex
\begin{algorithm}[ht]
\caption{Preprocessing}
\begin{small}
\label{alg2}
\begin{algorithmic}[1] 
\Function{\textnormal{preprocessAmbiguities}}{Image $I_\text{in}$}
    \For{\textbf{each} Point $\colorB{p}=(x,y)$ \textbf{in} $I_\text{in}$}
        \If{$I_\text{in}[\colorB{p}] > 0$ \textbf{and} \text{\textit{ambiguityMap}[\colorB{$p$}].size} $==$ 0}\label{alg2:isUnclustered}\Comment{Unprocessed pixel found}
            \State PointList \textit{neighbors} $=$ getDirectNeighbors($I_\text{in}$, \colorB{$p$})\label{alg2:getNbh}
            \If{\textit{neighbors}.size $>$ 2\ \textbf{or}\ containsFourCluster($I_\mathrm{in}$, \colorB{$p$})}\label{alg2:isCluster1}\Comment{Point $p$ is part of an ambiguity}
                \State \textbf{new} PointList \textit{clusterPoints}.append(\colorB{$p$})\label{alg2:list}\Comment{Append $p$ to new empty list}
                \State $c =$ 0
                \BeginBox[fill=gray, opacity=0.15]
                \While{$c$ $<$ \textit{clusterPoints}.size}\label{alg2:while}\Comment{Iteratively append points belonging to this ambiguity}
                    \State PointList \textit{neighbors} $=$ getDirectNeighbors($I_\text{in}$, \textit{clusterPoints}[$c$])
                    \For{\textbf{each} Point $p_n$ \textbf{in} \textit{neighbors}}
                        \If{$p_n$ \textbf{is not in} \textit{clusterPoints}}\label{alg2:notInClusterPoints} \Comment{Avoid double entries}
                            \State PointList \textit{neighbors} $=$ getDirectNeighbors($I_\text{in}$, $p_n$)
                            \If{\textit{neighbors}.size $>$ 2\ \textbf{or}\ containsFourCluster($I_\mathrm{in}$, $p_n$)}\label{alg2:isCluster2}
                                \State \textit{clusterPoints}.append($p_n$)
                            \EndIf    
                        \EndIf
                    \EndFor
                    \State $c=c+\text{1}$
                \EndWhile
                \EndBox
                \For{\textbf{each} Point $p_c$ \textbf{in} \textit{clusterPoints}}\label{alg2:writeAmbiguityMap}
                    \State \textit{ambiguityMap}[$p_c$] $=$ \textit{clusterPoints}\label{alg2:writeAmbiguityMapPoints}\Comment{Save \textit{clusterPoints} at each cluster point $p_c$}
                \EndFor
            \EndIf
        \EndIf
    \EndFor
\EndFunction
\end{algorithmic}
\end{small}
\end{algorithm}

%% file: 3c-pseudocode.tex
\begin{algorithm}[H]
\caption{Recursive Edge Tracing}
\begin{small}
\label{alg3}
\begin{algorithmic}[1] 
\State Initialize: \textit{edgeId} $=$ 0\label{alg3:initEdgeId}\Comment{Global counter incremented with each traced edge}
\Function{\textnormal{traceEdge}}{Image $I_\text{in}$, Point \colorB{$p$}, PointList \textit{\colorC{edge}}}
    \State \textit{\colorC{edge}}.append(\colorB{$p$})\label{alg3:edgeAppend}\Comment{Append point $p$ to current edge}
    \State \textit{edgeIdMap}[\colorB{$p$}].append(\textit{edgeId})\label{alg3:edgeIdMapAppend}
    \State PointList \textit{\colorA{neighbors}} $=$ getDirectNeighbors($I_\text{in}$, \colorB{$p$})\label{alg3:getDirectNeighbors}
    \State \textbf{new} PointList \textit{unvisitedNeighbors}
    \BeginBox[fill=gray, opacity=0.15]
    \If{\colorB{$p$} \textbf{is not in} ambiguity}\Comment{Initial exploration of direct neighbors}\label{alg3:part1Start}
        \For{\textbf{each} Point $p_n$ \textbf{in} \textit{\colorA{neighbors}}}
            \If{\textit{edgeIdMap}[$p_n$].size $==$ 0 \textbf{or} $p_n$ \textbf{is in} ambiguity}\label{alg3:part1Ambiguity}
                \State \textit{unvisitedNeighbors}.append($p_n$)\label{alg3:part1End}
            \EndIf
        \EndFor
    \EndIf
    \EndBox
    \BeginBox[fill=gray, opacity=0.15]
        \If{\textit{unvisitedNeighbors}.size $==$ 2}\label{alg3:start2Neighbors} \Comment{Further processing based on unvisited neighbors}
            \State \textbf{new} PointList \textit{edgePartOne}.append(\colorB{$p$}) \Comment{Create \textit{edgePartOne} and append point $p$}
            \State traceEdge($I_\text{in}$, \textit{unvisitedNeighbors}[0], \textit{edgePartOne})\label{alg3:recur1}
                    \State \textbf{new} PointList \textit{edgePartTwo}.append(\colorB{$p$}) \Comment{Create \textit{edgePartTwo} and append point $p$}
                    \State traceEdge($I_\text{in}$, \textit{unvisitedNeighbors}[1], \textit{edgePartTwo})\label{alg3:recur2}
                    \State mergeEdges($\textit{edgeId}-\text{2}$, $\textit{edgeId}-\text{1}$)\label{alg3:end2Neighbors}
        \ElsIf{\textit{unvisitedNeighbors}.size $==$ 1}\label{alg3:start1Neighbor}
            \State traceEdge($I_\text{in}$, \textit{unvisitedNeighbors}[0], \textit{\colorC{edge}}) \Comment{Continue with the only unvisited neighbor}\label{alg3:recur3}
        \ElsIf{\textit{unvisitedNeighbors}.size $==$ 0}\label{alg3:start0Neighbors}\Comment{Finish current edge}
            \State \textit{edgeList}.append(\textit{\colorC{edge}})
            \State $\textit{edgeId} = \textit{edgeId}+\text{1}$ \label{alg3:increaseEdgeId1}\label{alg3:end0Neighbors}
        \EndIf
    \EndBox
\EndFunction
\end{algorithmic}
\end{small}
\end{algorithm}
%




%% file: 4-evaluation.tex
\section{Evaluation}
\label{sec:evaluation}

\begin{figure}[t]
\centering %
\vspace{0.5cm}
\includegraphics{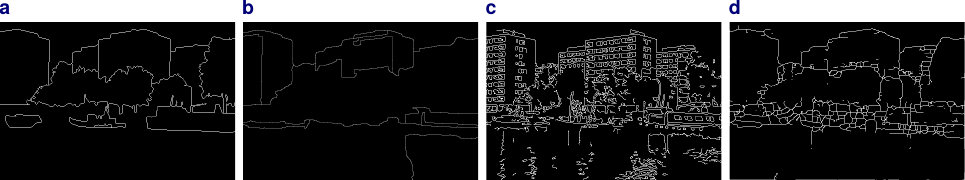}
\parbox{\textwidth}{\raggedright
{\scriptsize
\vspace{-0.1cm}
\hspace{0.75cm} Manual annotation \cite{arbelaez2011}
\hspace{1.8cm} gPb-owt-ucm \cite{arbelaez2011}
\hspace{2.5cm} Canny \cite{canny1986}
\hspace{2.8cm} HED \cite{xie2015}
}}
\vspace{-0.2cm}
\caption{Example test image from our dataset for the evaluation based on image 78098 from the BSDS500 dataset. \subref{a} Manual binary annotation provided with the BSDS500 dataset. \subref{b} Result of the gPb-owt-ucm method provided with the BSDS500 dataset (scaled). \subref{c} Canny edge detector result. \subref{d} HED edge detector result with an additional thinning step for binarization. See text for further details.}
\label{fig:dataset}
\end{figure}

In \cref{fig:comparison}, we have already compared CCL, MNT, and FCM with our method. For further insights into the corresponding component decomposition, redundancy, and missing pixels, we have created a dataset with binary edge images based on the BSDS500 dataset \cite{arbelaez2011} and analyzed the results.

\subsection{Dataset Construction}
\label{sec:dataset}
For each of the 500 images from the BSDS500 dataset, we have created four binary edge images. An example is shown in \cref{fig:dataset}. For this purpose, we have used one of the manual binary annotations and the result of the gPb-owt-ucm method, both provided with the dataset. Furthermore, we have applied the Canny edge detector \cite{canny1986} and the HED method \cite{xie2015} to each image. Since multiple manual annotations are provided (five on average per image), we randomly selected one per image. The images from the gPb-owt-ucm method have twice the dimensions of the original input images and the other binary edge images ($481 \times 321$\,px vs.\ $963 \times 643$\,px). The Canny edge detector directly provides binary edge images through its included non-maximum suppression and hysteresis thresholding. Since the HED method produces broad, non-binary edges, we applied the Zhang-Suen thinning algorithm \cite{zhang1984} to obtain the corresponding binary edge images. As gPb-owt-ucm is a classical learning-based method, Canny a traditional gradient-based method, and HED is a deep learning-based method, a representative range of edge detection methods is covered. The following evaluation compares the different methods from a general perspective without focusing on specific tasks. Which method is best generally depends on the task.

\subsection{Method Characteristics Overview}
\label{sec:characteristics}
CCL determines one component for each set of connected pixels so that each pixel from the input image belongs to only one connected component. The method does not lead to any missing pixels. In the following evaluation, we use CCL as the baseline method, as it provides information about the structure and complexity of the input image. In \cref{fig:comparison}\sub{b}, for example, CCL identifies only one connected component.

\input{4a-tables}
\input{4b-tables}

MNT determines edges by following around the outer boundary of each connected component and therefore leads to one edge per component. In comparison, FCM and our method can segment each component into several edges (and ambiguities). MNT can trace the same pixel twice in the same edge and lead to missing pixels, as shown in \cref{fig:comparison}\sub{c}.

FCM can determine several edges for each connected component and also detects hierarchies between edges (parents and childs), where the parent edges are essentially identical to the MNT results. FCM can trace the same pixel twice in the same edge, trace the same pixel in different edges, and also lead to missing pixels, as shown in \cref{fig:comparison}\sub{d}.

Our method can determine several edges and ambiguities for each connected component. In our method, the same pixel can be traced in different edges (in that case, it is an ambiguity point), but unlike MNT and FCM, not in the same edge. Our method does not lead to any missing pixels, as each pixel is either part of an edge or ambiguity (or both), as shown in \cref{fig:comparison}\sub{e}.

\subsection{Method Analysis}
\label{sec:method-analysis}
In Tables \ref{tab:average-pixels}--\ref{tab:px-to-seg}, we refer to segments as the main entities extracted by each method. In CCL, one segment corresponds to one connected component, whereas in MNT, FCM, and ours, to one edge. In our method, we consider ambiguity points without edges separately and not as segments. Tables \ref{tab:average-pixels}--\ref{tab:traced-vs-edge-px} include the results for \cref{fig:comparison} to facilitate the interpretation of the data presented.

\cref{tab:average-pixels} shows the average number of pixels per segment for each method across the different edge detectors. To compute the values, we have counted the total number of segment pixels and divided by the number of segments. In \cref{fig:comparison}, for example, our method has captured 52 pixels in 9 segments ($52/9 \approx 5.8$). MNT leads to the highest values as it traces certain pixels twice per segment (edge). In summary, our method provides the smallest segments and therefore the most detailed breakdown into edges.

\cref{tab:segments} confirms these results from another perspective. Here, we have counted the total number of connected components and divided by the number of segments. In \cref{fig:comparison}, for example, our method has traced 9 edges for the given component. CCL and MNT naturally have a value of $1.00$, since both methods provide exactly one segment per component.

\cref{tab:traced-vs-edge-px} shows the total number of segment pixels in relation to the total number of edge pixels (pixels labeled 1 in the input image). In \cref{fig:comparison}, for example, our method has captured 52 segment pixels and the input image contains 47 edge pixels ($52/47 \approx 1.11$). We interpret the values as a measure for redundancy and missing pixels, where smaller values mean lower redundancy and more missing pixels. In general, good values are close to $1.00$. MNT has missed certain pixels so that the value is even smaller than $1.00$. In summary, our method provides the lowest redundancy.

\cref{tab:px-to-seg} shows the number of segments to which each segment pixel has been assigned (on average over the entire dataset). In \cref{fig:comparison}, for example, our method has traced 52 segment pixels, and 49 of them are in only one edge ($49/52 \approx 94.23\,\%$; not shown in the table). The ``--'' indicate that CCL and MNT cannot assign pixels to more than one element. The proportion assigned to zero segments in our method corresponds to the number of ambiguity points without edges. Note that our method does still capture these points; they simply do not contain any edges. In summary, our method assigns most pixels to a single segment.
\enlargethispage{\baselineskip}

\subsection{Runtime Analysis}
\label{sec:runtime-analysis}
\begin{figure}[t]
\centering
\includegraphics[scale=1.0]{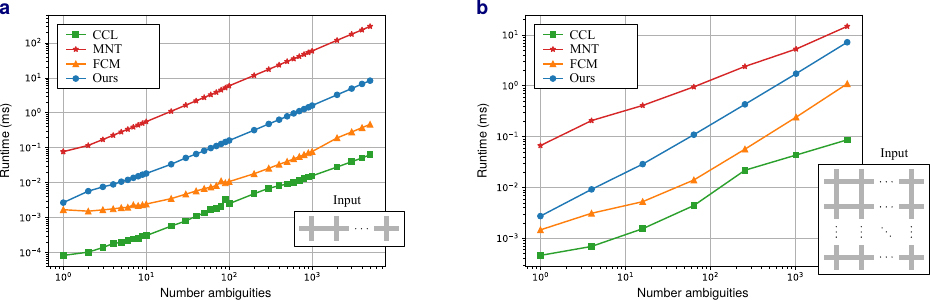}
\caption{Runtime analysis of the different methods for an increasing number of ambiguities (connected $\text{5}\times\text{5}$ px crosses, cf.\ bottom right of the subfigures). CCL, FCM, and our method have been implemented in C++, and MNT in Python. \subref{a} Increasing number of ambiguities in a row. \subref{b} Increasing number of ambiguities in a square pattern.}
\label{fig:runtime}
\end{figure}

To analyze the runtime of the different methods, we have created regular patterns with an increasing number of ambiguities, and calculated the mean execution times over 10,000 runs for each method (Processor: Intel i7-13700K). The results indicate an approximately linear relationship for all methods, both for an arrangement in a row (\cref{fig:runtime}\sub{a}) and in a square pattern (\cref{fig:runtime}\sub{b}). Concerning our method, this is plausible because each additional ambiguity adds the same number of operations, both during the preprocessing and edge tracing. Note that the time is measured in milliseconds. Processing a typical dataset image with our method takes $< 2\, \text{ms}$, it is therefore real-time capable. Depending on the edge detection method, there are approximately 50 (Manual annotation) to 500 (Canny) ambiguities in a dataset image.

%% file: 4a-tables.tex
\begin{table}[t]
\captionsetup{justification=raggedright, singlelinecheck=false}
\begin{minipage}{0.53\textwidth}
\small 
\begin{tabular}{lcccccc}
\toprule
\multirow{2}{*}{Method\vspace{-0.1cm}} & \multicolumn{5}{c}{Avrg.\ No.\ Pixels per Segment (px)} \\
\cmidrule(lr){2-6}
                        & {Fig.\ \ref{fig:comparison}} & {Anno} & {gPb} & {Canny} & {HED} \\
\midrule
CCL                     & 47.0           & 597.2 & 2178.7 & 46.9 & \phantom{0}95.8 \\
MNT                     & 46.0           & 736.1 & 2441.0 & 69.1 & 114.9 \\
FCM                     & 22.0           & 208.1 & \phantom{0}422.0 & 58.0 & \phantom{0}86.5 \\
Ours & \phantom{0}\textbf{5.8} & \phantom{0}\textbf{39.6} & \phantom{00}\textbf{78.3} & \phantom{0}\textbf{9.9} & \phantom{0}\textbf{15.8}\\
\bottomrule
\end{tabular}
\captionsetup{type=table}
\caption{Pixels per segment, smallest values bold.}
\label{tab:average-pixels}
\end{minipage}
\begin{minipage}{0.46\textwidth}
\small 
\vspace{0.025cm}
\begin{tabular}{lcccccc}
\toprule
\multirow{2}{*}{Method\vspace{-0.1cm}} & \multicolumn{5}{c}{Avrg.\ No.\ Segments per CC} \\
\cmidrule(lr){2-6}
                        & {Fig.\ \ref{fig:comparison}} & {Anno} & {gPb} & {Canny} & {HED} \\
\midrule
CCL                     & 1.00 & \phantom{0}1.00 & \phantom{0}1.00 & 1.00 & 1.00 \\
MNT                     & 1.00 & \phantom{0}1.00 & \phantom{0}1.00 & 1.00 & 1.00 \\
FCM                     & 3.00 & \phantom{0}5.75 & \phantom{0}9.25 & 1.29 & 2.01 \\
Ours                    & \textbf{9.00} & \textbf{15.47} & \textbf{28.27} & \textbf{5.12} & \textbf{6.46} \\
\bottomrule
\end{tabular}
\captionsetup{type=table}
\caption{Segments per CC, highest values bold.}
\label{tab:segments}
\end{minipage}
\end{table}

%% file: 4b-tables.tex
\begin{table}[t]
\captionsetup{justification=raggedright, singlelinecheck=false}
\begin{minipage}{0.53\textwidth}
\small 
\vspace{0.3cm}
\begin{tabular}{lcccccc}
\toprule
\multirow{2}{*}{Method\vspace{-0.1cm}} & \multicolumn{5}{c}{No.\ Segment Pixels vs.\ Edge Pixels} \\
\cmidrule(lr){2-6}
                        & {Fig.\ \ref{fig:comparison}} & {Anno} & {gPb} & {Canny} & {HED} \\
\midrule
CCL                     & 1.00 & \phantom{0}1.00\phantom{0} & \phantom{0}1.00\phantom{0} & 1.00 & 1.00 \\
MNT                     & \textbf{0.98} & 1.23 & 1.12 & 1.47 & 1.20 \\
FCM                     & 1.41 & 2.00 & 1.79 & 1.59 & 1.81 \\
Ours                    & 1.11 & \textbf{1.03} & \textbf{1.02} & \textbf{1.08} & \textbf{1.06} \\
\bottomrule
\end{tabular}
\captionsetup{type=table}
\caption{Segment vs.\ edge pixels, closest to CCL bold.}
\label{tab:traced-vs-edge-px}
\end{minipage}
\begin{minipage}{0.46\textwidth}
\small 
\vspace{0.29cm}
\begin{tabular}{lccccccc}
\toprule
\multirow{2}{*}{Method\vspace{-0.1cm}} & \multicolumn{5}{c}{Pixel to Segment Assignment (\%)} \\
\cmidrule(lr){2-6}
                        & 0 & 1 & 2 & 3 & >3 \\
\midrule
CCL                     & \phantom{0}0.00  & 100.00 & --    & --   & -- \\
MNT                     & 14.26 & \phantom{0}85.74  & --    & --   & -- \\
FCM                     & \phantom{0}0.03  & \phantom{0}68.31  & 31.56 & 0.11 & 0.00 \\
Ours                    & \phantom{0}0.33  & \phantom{0}96.95  & \phantom{0}0.31  & 2.40 & 0.02 \\
\bottomrule
\end{tabular}
\captionsetup{type=table}
\caption{Assignment of pixels to how many segments.}
\label{tab:px-to-seg}
\end{minipage}
\end{table}

%% file: 5-conclusion.tex
\section{Conclusions and Final Remarks}
\label{sec:conclusion}
Despite using only four straightforward principles, our ambiguity model can handle complex structures in binary edge images in an intuitive and effective manner. With the specialized design of our method, ambiguities can be resolved in a more direct manner compared to existing methods (without using graphs, etc.). We have found that our model is a natural extension of the concept of single-pixel ambiguities. Our implementation is divided into a preprocessing and an edge tracing part, which provides clear operational control and helped us to formulate a proof of correctness. We have shown that our method can effectively resolve complex ambiguities in different application examples. Compared to others, our method provides the most detailed breakdown of binary edge images into segments while also reducing redundancy (double reading of edge pixels). While our method can also process regions, it is not intended for this purpose and simply identifies them as coherent ambiguities. In future works, our method could be applied to different tasks on larger datasets.

A rather theoretical limitation of our method is the possible depth of recursion (which is not specific to our method, but a general characteristic of recursions). Since recursion is only used for the edge tracing, the maximum depth to be handled corresponds to the maximum edge length.